# AgentWeave: Routing Before Reasoning for Efficient Function Calling in Tool-Rich Language Models

Saurav Singla • Aarav Singla • Advik Gupta • Parnika Gupta

Open-source research project: github.com/sauravsingla/agentweave

## Abstract

Large language models increasingly operate over large collections of tools, functions, APIs, and specialized agents. As the candidate action space grows, a function-calling model must process more schemas, consume more prompt tokens, and distinguish among increasingly similar or irrelevant alternatives. We study a complementary systems strategy: reduce the candidate set before language-model inference while leaving the downstream model unchanged. We introduce AgentWeave, a deterministic pre-inference routing layer that constructs a bounded model-visible action space using eligibility, requirement, capability, and routing signals. We evaluate AgentWeave with a frozen BFCL-derived routing-pressure protocol using the public MadeAgents/Hammer2.1-1.5b model. On 48 fresh BFCL V4 multiple-function tasks, AgentWeave achieves 6/48 (12.5%) native BFCL successes, whereas all-tools, deterministic random top-8, and semantic top-8 baselines each achieve 0/48. The paired success difference is +12.5 percentage points with a 10,000-resample paired bootstrap 95% confidence interval of +4.17 to +22.92 points and exact McNemar p=0.03125. Relative to all-tools exposure, AgentWeave presents 70.18% fewer tools, uses 61.70% fewer input tokens, and exhibits 50.95% lower mean local-model latency. The result is deliberately narrow: this is a BFCL-derived routing-pressure study rather than an official full BFCL leaderboard score, and absolute task success remains low. The evidence nevertheless shows that candidate-space construction can materially affect a fixed model's function-calling behavior and motivates evaluating routing as a distinct stage before model reasoning.

## 1. Introduction

Function calling has become a central interface between language models and external computation. Modern systems use model-generated calls to search services, databases, software libraries, enterprise APIs, web actions, and domain-specific tools. Toolformer [1], ReAct [2], Gorilla [3], API-Bank [4], ToolLLM [5], Hammer [6], ToolACE [7], and the Berkeley Function Calling Leaderboard (BFCL) [8] demonstrate the breadth of this research direction. Yet most evaluation pipelines still treat the candidate tool catalog as a fixed input to the model. This is convenient for benchmarking, but it obscures a systems variable that becomes important as catalogs grow: which functions should the model see at all?

The underlying hypothesis of this paper is simple. Function calling can be decomposed into candidate-space construction followed by language-model reasoning. An application may first remove tools that are unavailable because of role, tenant, policy, deployment, or capability constraints. A task-aware router can then reduce the remaining catalog further. The language model subsequently chooses among a smaller set. This decomposition does not require fine-tuning the downstream model and can therefore be evaluated using controlled same-model comparisons.

- formalize pre-inference tool routing as a distinct stage in function calling;
- describe AgentWeave as a deterministic, model-independent routing layer;
- present a frozen 48-task BFCL-derived replication with matched baselines;
- quantify native success, candidate retention, token use, and latency; and
- state explicit evidence boundaries, limitations, and reproducibility controls.

## 2. Background and Related Work

Tool-augmented language modeling has evolved along several complementary axes. Toolformer [1] showed that a language model can learn when and how to invoke external APIs using self-supervised training. ReAct [2] integrated reasoning and acting in an interleaved trajectory, emphasizing that external actions and internal reasoning can reinforce each other. Gorilla [3] focused on API invocation at scale and demonstrated the value of retrieval when API documentation changes. These systems established that tool use is not merely output formatting; it is a structured decision problem under external constraints.

Benchmarking work subsequently broadened the evaluation surface. API-Bank [4] evaluates planning, retrieval, and tool invocation across runnable APIs. ToolLLM [5] introduced ToolBench, a large collection of real-world APIs with instruction and solution-path annotations, and equipped ToolLLaMA with neural API retrieval. BFCL [8] developed an executable and AST-based evaluation framework spanning single, parallel, multiple, live, and agentic settings. The current BFCL V4 further incorporates multi-turn and agentic evaluations, while retaining single-turn categories that are useful for controlled function-selection studies.

Another line of work improves the function-calling model itself. Hammer [6] targets lightweight on-device function calling and uses function masking and irrelevance-focused data to increase robustness. ToolACE [7] constructs large, verified function-calling training corpora and demonstrates that carefully synthesized data can produce strong function-call performance. These approaches improve the model; AgentWeave instead holds the model fixed and changes only the candidate functions visible before inference.

### 2.1 Candidate Retrieval Versus Final Selection

Tool retrieval and final function selection are related but not identical. A retrieval module can succeed when the required function is present in the returned set even if it is not ranked first. The downstream language model is then responsible for choosing the final function and generating arguments. This distinction suggests that retrieval-stage metrics such as Recall@K, compression, and retained-required-function rate should be reported separately from native function-call accuracy.

This separation is particularly important for multiple-function tasks. A query may require a combination of functions; retaining one relevant function is insufficient if another required function is removed. Consequently, all-candidates-retained rate complements average recall. At the same time, retaining every potentially relevant function may recreate the original schema-competition problem. The appropriate routing target is therefore a high-recall but bounded candidate set, not necessarily a single predicted function.

### 2.2 Positioning of AgentWeave

AgentWeave is best viewed as an orchestration layer that can be placed between a source catalog and an existing model. It does not propose a replacement language model, new decoding algorithm, or modified benchmark evaluator. In this study, the scientific question is deliberately restricted to pre-inference routing: can the same function-calling model behave differently when the candidate set is constructed by a deterministic routing layer rather than exposed in full or reduced by simpler baselines?

The resulting contribution is therefore systems-oriented but directly relevant to computation and language. Function schemas are textual structures consumed by the model; candidate composition changes the language context over which the model reasons. The experiment asks whether selective context construction can improve function-calling behavior while simultaneously reducing the amount of schema text the model must process.

## 3. Problem Formulation and Architecture

Let x denote a natural-language request, F={f1,...,fn} an available function catalog, and M a fixed function-calling language model. A conventional system predicts y=M(x,F). We introduce a routing function R that maps the request and source catalog to a subset F'=R(x,F), with F' contained in F. The downstream prediction becomes y=M(x,F'). The model, model weights, generation settings, and evaluation logic remain unchanged across conditions.

Routing must balance compression against retention. If |F'| is too large, the model still processes most of the source catalog and gains little efficiency. If |F'| is too small, required functions may be removed before inference. For a task with required set G contained in F, a useful router should maximize retention of G while minimizing |F'| and preserving a candidate composition that is tractable for the downstream model.

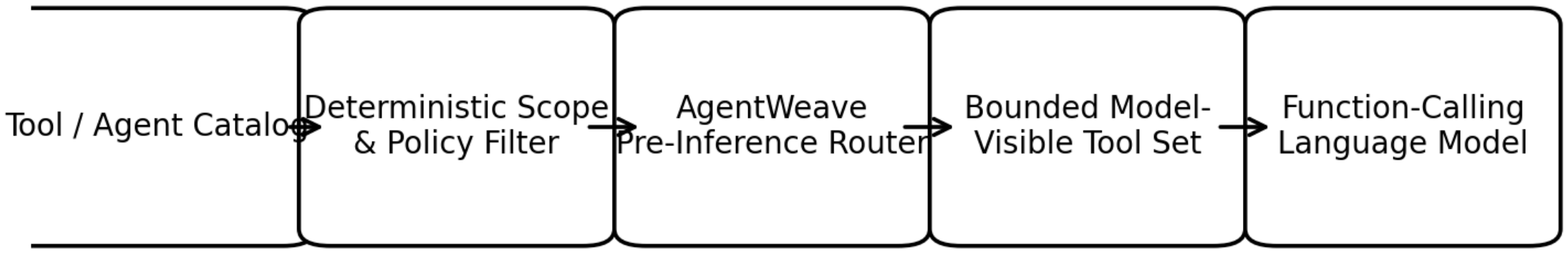

*Figure 1. AgentWeave separates candidate-space construction from downstream model reasoning and execution authorization.*

### 3.1 Deterministic Scope and Policy Filtering

Production tool catalogs often contain candidates that can be excluded without semantic interpretation. Role, tenant, permission, deployment environment, jurisdiction, product surface, or capability tags may deterministically restrict the eligible set. AgentWeave's emerging architecture therefore treats policy-first scope filtering as a separate upstream stage. Dynamic routing is optional: if deterministic filtering already yields a sufficiently small catalog, no additional semantic routing is necessary.

### 3.2 Requirement-Aware Routing

For the remaining candidates, AgentWeave derives requirement signals from the request and matches them against candidate descriptions, capabilities, and provider groupings. The router applies a bounded budget so that only a small subset is exposed to the language model. The design is intentionally model-independent: local models, hosted models, or provider-neutral adapters can consume the selected function list without changing the router interface.

### 3.3 Provenance and Failure Localization

A central practical benefit of separating stages is failure localization. If the final tool call is wrong, a developer should be able to determine whether the correct tool was filtered out, remained visible but was not selected, was selected but denied by authorization, or executed incorrectly. AgentWeave therefore motivates stage-level provenance containing source-catalog identity, routing configuration, candidate counts, filtered candidates, model-visible candidates, selected function, and execution outcome. The portable artifact should record decisions and provenance rather than private reasoning traces.

This decomposition also clarifies security boundaries. Pre-model filtering can reduce exposure to irrelevant or malicious tool descriptions, but it is not a substitute for authorization. A deterministic fail-closed authorization gate should remain after model selection and before execution. In other words, routing optimizes what the model sees; authorization governs what the system is allowed to do.

## 4. Experimental Design

### 4.1 Benchmark Source and Task Selection

The primary experiment is the frozen BFCL routing-pressure v6 replication recorded in the AgentWeave repository. The study uses 48 fresh BFCL V4 multiple-function tasks with zero overlap with the earlier 12-task v5 pilot. BFCL evaluates function calling across realistic candidate functions and provides native evaluation machinery; the multiple category is particularly suitable because correct behavior can require selecting among several candidate functions.

Task selection is content-blind and frozen before scoring. The study records the exact scored source revision, BFCL/Gorilla revision, workflow run, artifact identifier, and artifact digest. After the first successful scored run, the experiment is treated as immutable. Any outcome-affecting modification requires a new study identifier and a fresh untouched sample. This rule is intended to reduce post-hoc tuning against observed benchmark outcomes.

### 4.2 Fixed Downstream Model

All conditions use MadeAgents/Hammer2.1-1.5b, a lightweight function-calling model related to the Hammer family [6]. The model is held fixed across all routing strategies. The evaluation is local and keyless and incurs no external API spend. This same-model design isolates candidate-presentation effects from model scaling or provider differences.

### 4.3 Routing-Pressure Construction

Each task is placed in a deterministic 16-tool pressure environment before strategy-specific selection. The original BFCL question is not rewritten and native BFCL evaluation is preserved after restoring the expected evaluation structure. The intervention therefore operates on the model-visible candidate context rather than on benchmark ground truth.

| Condition | Selection rule | Nominal budget |
|---|---|---|
| All tools | Expose full pressure set | 16 |
| Random top-8 | Deterministic random reduction | 8 |
| Semantic top-8 | Embedding/similarity reduction | 8 |
| AgentWeave | Requirement/capability-aware bounded routing | up to routed budget |

*Table 1. Matched routing conditions used in the v6 study.*

### 4.4 Metrics

The primary outcome is native BFCL success. Efficiency metrics include mean tools shown, aggregate model-input tokens, and mean local-model latency. Routing diagnostics include mean original-candidate recall and the fraction of tasks for which all original candidates are retained. Statistical analysis uses paired comparisons because each strategy is evaluated on the same tasks.

### 4.5 Statistical Analysis

For the v6 replication, the paired native-success difference between AgentWeave and each matched baseline is summarized with a 10,000-resample paired bootstrap 95% confidence interval. Exact McNemar tests assess the significance of discordant paired binary outcomes. These tests are appropriate for the matched task-level design and avoid treating strategies as independent samples.

The study reports both positive and negative results. Candidate retention is not optimized after seeing the final score, and semantic retrieval is allowed to outperform AgentWeave on retention diagnostics. This transparency matters because a routing method can appear stronger if only favorable metrics are retained.

## 5. Main Results

### 5.1 Native Function-Calling Success

| Strategy | Native success | Mean tools | Input tokens | Mean latency |
|---|---|---|---|---|
| All tools | 0/48 (0.0%) | 16.00 | 124,821 | 55.92 s |
| Random top-8 | 0/48 (0.0%) | 8.00 | 69,790 | 30.56 s |
| Semantic top-8 | 0/48 (0.0%) | 8.00 | 70,271 | 35.87 s |
| AgentWeave | 6/48 (12.5%) | 4.77 | 47,805 | 27.43 s |

*Table 2. Frozen BFCL-derived routing-pressure v6 outcomes and efficiency measurements.*

AgentWeave is the only strategy to produce any native BFCL successes in the 48-task replication, reaching 6/48=12.5%. Each matched baseline yields 0/48. The paired AgentWeave advantage over each baseline is therefore +12.5 percentage points. The 10,000-resample paired bootstrap 95% confidence interval is +4.17 to +22.92 percentage points, and the exact McNemar test gives $p=0.03125$.

The absolute success rate is low and should not be obscured by the relative comparison. The experiment is deliberately difficult: all conditions use the same 1.5B model under a 16-tool pressure setting, and three reasonable alternatives fail on every task. The evidence supports a limited conclusion: under this tested setting, changing candidate-space construction alters the downstream behavior of a fixed function-calling model.

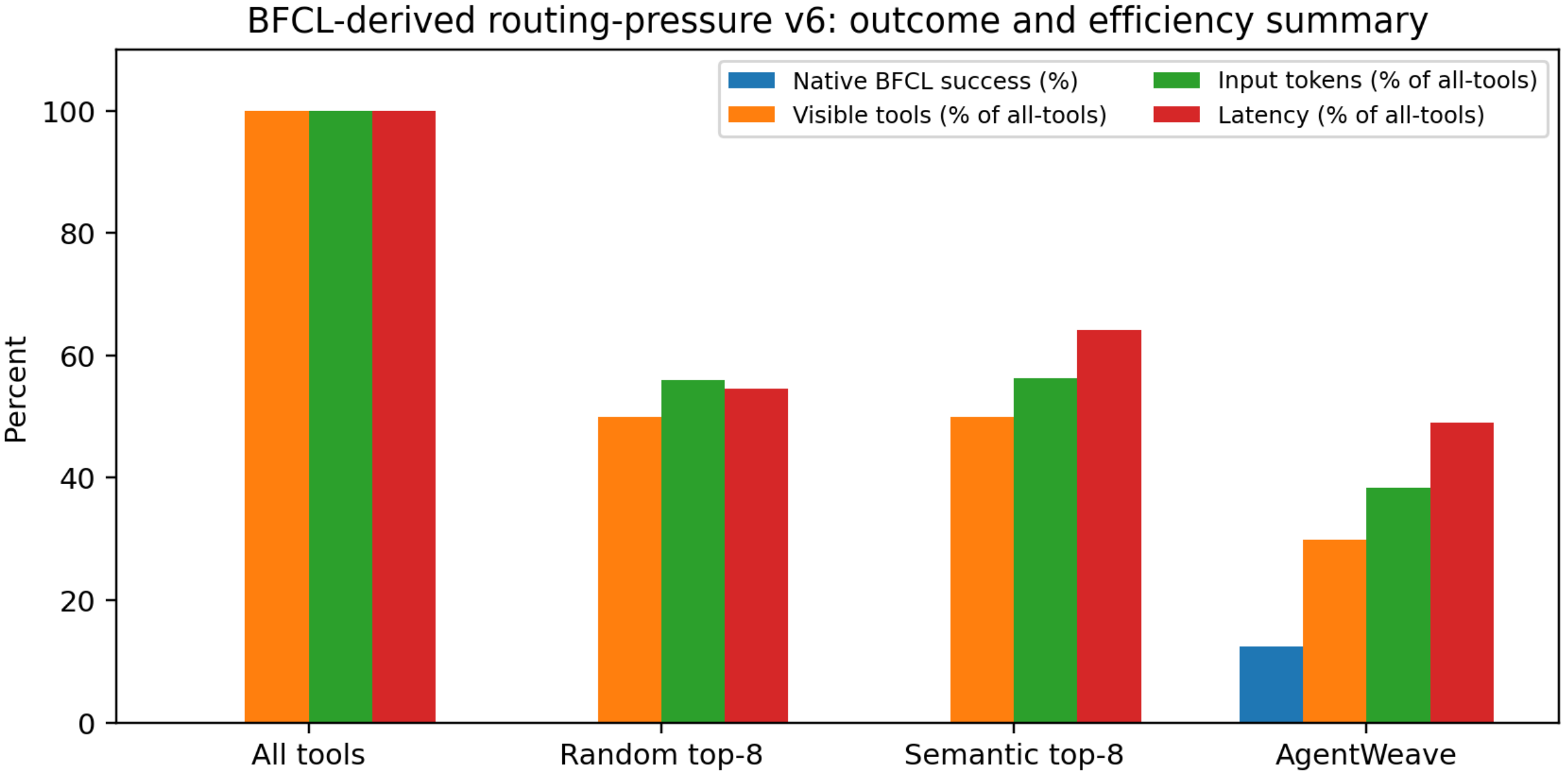

*Figure 2. Native success and normalized efficiency measures for the four v6 conditions. Efficiency bars are expressed relative to all-tools exposure.*

### 5.2 Efficiency Effects

Against all-tools exposure, AgentWeave shows 70.18% fewer model-visible tools, 61.70% fewer input tokens, and 50.95% lower mean local-model latency. The reductions are directionally consistent with the intended role of pre-inference routing: shorter schema context should generally require fewer tokens and less model processing time. However, latency is hardware- and implementation-dependent and should not be interpreted as a universal speedup.

The random and semantic top-8 baselines also reduce token count and latency, confirming that some efficiency gain follows directly from showing fewer tools. AgentWeave's contribution is therefore not the existence of compression itself; it is the combination of stronger observed native success in this replication with a smaller average visible tool set.

## 6. Candidate Retention and Failure Analysis

### 6.1 Retention Diagnostics

| Strategy | Mean original-candidate recall | All original candidates retained |
|---|---|---|
| All tools | 100.00% | 100.00% |
| Random top-8 | 48.44% | 10.42% |
| Semantic top-8 | 86.81% | 66.67% |
| AgentWeave | 76.91% | 54.17% |

*Table 3. Routing diagnostics reveal that higher candidate recall alone does not explain native success.*

Semantic top-8 has the highest candidate-retention diagnostics among the compressed conditions: 86.81% mean original-candidate recall and 66.67% all-candidates-retained rate. AgentWeave is lower on both measures, at 76.91% and 54.17%, respectively. Yet semantic top-8 achieves 0/48 native successes whereas AgentWeave achieves 6/48.

This result is important because it prevents a simplistic interpretation of the experiment. AgentWeave does not win merely because it retains more of the original BFCL candidates. It does not. Instead, the observed successes may depend on the composition and competition structure of the selected set. A model that sees the required function together with several close distractors may behave differently from the same model seeing the required function in a more discriminative set.

### 6.2 Failure Taxonomy

- Missing required candidate: at least one function needed by the benchmark is removed before inference.
- Partial multiple-function coverage: some, but not all, required functions survive routing.
- Schema competition: all required functions remain visible, but closely related distractors cause incorrect selection.
- Argument-generation error: the correct function is selected, but arguments or formatting fail native evaluation.
- Downstream model failure: the candidate set is adequate, but the model does not produce a valid call.
- Over-compression: routing saves context but reduces required-function recall below a useful operating point.

### 6.3 Why Recall@K Is a Better Routing Objective Than Hit@1

The router is not necessarily the final decision maker. If a source catalog contains 500 functions and the required function is retained in a high-quality set of six, the router has performed a useful service even if that function is ranked second within the six. The language model is explicitly retained as the final function selector. For this architecture, required-tool Recall@K and all-required-tools-retained rate are therefore more faithful routing objectives than router Hit@1 alone.

A future AgentWeave evaluation should report an operating curve over candidate budget K: retention versus compression, followed by downstream native success. Such a curve would reveal whether improvements come from better retrieval, from a more favorable candidate composition, or from a larger budget that simply approaches the all-tools condition. It would also permit policy-driven selection of an operating point according to latency, token budget, or risk tolerance.

The v6 evidence indicates that candidate retention and native success can diverge. This motivates a two-stage research agenda: first optimize retrieval for high recall under bounded context; then study how the downstream model's reasoning changes as distractor composition, schema similarity, and candidate ordering vary.

## 7. Replication, Robustness, and Scientific Boundaries

### 7.1 Relationship to the v5 Pilot

The preceding frozen v5 pilot evaluated 12 tasks and produced 2/12=16.67% native success for AgentWeave versus 0/12 for all three matched baselines. Because only two discordant successes were observed, exact McNemar p=0.5 and the sample was insufficient for a strong statistical claim. The v6 study was preregistered as a larger untouched replication with 48 fresh tasks and zero overlap with v5.

The v6 outcome, 6/48=12.5% versus 0/48 for each baseline, preserves the directional finding of the pilot while increasing the number of discordant successful cases. The experimental model, routing budgets, baseline definitions, 16-tool pressure setting, generation settings, and native BFCL evaluation were held constant between v5 and v6. This relationship is stronger evidence than repeatedly modifying the router on the same observed benchmark split.

### 7.2 Frozen Evidence and Anti-Tuning Controls

AgentWeave maintains a broader research practice in which scored studies are frozen after first successful evaluation and subsequent router versions are tested on newly introduced holdouts. The repository also retains weak and negative results. For example, earlier frozen routing evaluations show that simple embedding baselines can outperform the original router on already-observed data. Retaining these results reduces the temptation to present only favorable comparisons and makes the evolution of the routing system auditable.

### 7.3 What the Study Does Not Claim

- It is not an official full BFCL leaderboard score.
- It does not establish that AgentWeave is universally superior to semantic retrieval.
- It does not establish production-scale end-to-end task success.
- It does not establish a universal latency or token reduction across hardware and model providers.
- It does not show that routing alone solves authorization, security, or execution correctness.
- It does not show generalization across multiple function-calling models.

### 7.4 Why the Negative Results Matter

The semantic baseline's stronger retention but zero native success is scientifically useful even though it complicates a clean narrative. Likewise, AgentWeave's absolute 12.5% success rate is low. Reporting these facts constrains the interpretation to what the experiment demonstrates. The central contribution is not a state-of-the-art leaderboard claim but evidence that the candidate set is an experimentally meaningful variable for a fixed model.

This distinction aligns with arXiv's role as a research-sharing platform rather than a venue that guarantees peer-reviewed correctness. A responsible preprint should make experimental boundaries explicit, provide enough detail for reproduction, and avoid presenting derived routing-pressure numbers as official benchmark rankings.

## 8. Deep Result Analysis and Mechanistic Interpretation

The v6 experiment is most informative when the four outcomes are interpreted jointly rather than as an accuracy-only comparison. AgentWeave changes three quantities at once: the size of the model-visible action space, the amount of schema text entering the model, and the composition of competing functions. Native BFCL success then measures whether this altered context is sufficient for the fixed Hammer2.1-1.5b model to produce an evaluator-accepted call. Because model weights and generation settings are held constant, differences across conditions are attributable to candidate presentation within this protocol, although the experiment does not isolate which individual routing feature is causal.

### 8.1 Effect Size in the Matched 48-Task Replication

AgentWeave succeeds on 6 of 48 tasks while each comparator succeeds on 0 of 48. The absolute risk difference is therefore 0.125. In practical terms, the routed condition converts six matched tasks from baseline failure to native success without an observed success in the opposite direction against these zero-success baselines. This asymmetry is why a paired test is preferable to comparing unrelated proportions. The reported exact McNemar p=0.03125 evaluates discordant outcomes on the same tasks, while the 10,000-resample paired bootstrap interval of +4.17 to +22.92 percentage points quantifies uncertainty in the task-sampled improvement.

The result is evidence of an effect in this frozen routing-pressure setting, not a population-wide estimate of AgentWeave accuracy. With 48 tasks, a small number of additional successes or failures would materially change the observed rate. Statistical significance strengthens the claim that candidate presentation mattered in this matched sample, but does not establish generalization to other BFCL categories, models, catalog sizes, or routing budgets.

| Quantity | Observed result | Interpretation |
|---|---|---|
| Native success difference | +12.5 percentage points | 6 routed successes vs. 0 |
| Paired bootstrap 95% CI | +4.17 to +22.92 points | Task-sampled uncertainty |
| Exact McNemar | p = 0.03125 | Matched binary outcome test |
| Absolute success | 6/48 (12.5%) | Still low; preliminary |

*Table 4. Statistical interpretation of the frozen v6 result.*

### 8.2 Compression, Token Economy, and Latency

The efficiency evidence is coherent across three measurements. AgentWeave reduces the average visible catalog from 16.00 to 4.77 tools, a reduction of 11.23 tools per task or 70.18%. Aggregate input tokens fall from 124,821 to 47,805, a reduction of 77,016 tokens or 61.70%. Mean local-model latency falls from 55.92 s to 27.43 s, a reduction of 28.49 s or 50.95%. All three quantities move in the direction expected if tool-schema exposure is a meaningful variable component of inference cost.

The reduction is not proportional across metrics. AgentWeave exposes about 29.8% of the all-tools function count, yet retains about 38.3% of the input-token volume and 49.1% of the mean latency. Fixed request text, system instructions, tokenizer and generation overhead, and other implementation costs remain even after schemas are removed. Tool routing therefore reduces a variable component of inference rather than scaling total request cost linearly.

| Metric | All tools | AgentWeave | Absolute change | Reduction |
|---|---|---|---|---|
| Mean tools | 16.00 | 4.77 | -11.23 | 70.18% |
| Input tokens | 124,821 | 47,805 | -77,016 | 61.70% |
| Mean latency | 55.92 s | 27.43 s | -28.49 s | 50.95% |

*Table 5. Absolute and relative efficiency changes against full exposure.*

### 8.3 The Candidate-Composition Paradox

The strongest counter-result is that semantic top-8 preserves more of the original benchmark catalog than AgentWeave. Its mean original-candidate recall is 86.81% versus 76.91%, and it retains all original candidates on 66.67% of tasks versus 54.17% for AgentWeave. Yet semantic top-8 produces 0/48 native successes. The AgentWeave advantage therefore cannot be explained by a simple 'more benchmark candidates retained' story; it occurs despite weaker retention on these diagnostics.

A plausible mechanism is distractor geometry. Two candidate sets can both contain a required function while presenting very different decision surfaces to a small language model. Semantic retrieval may collect near-neighbor functions whose names, descriptions, or argument schemas compete with the target. Requirement/capability-aware routing may retain fewer original candidates while eliminating some confusing alternatives. The frozen artifacts do not directly measure pairwise schema similarity for successful versus failed tasks, so this remains a testable hypothesis rather than a demonstrated causal mechanism.

This motivates a targeted follow-up: measure required-function retention, distractor similarity, schema redundancy, candidate ordering, and native success for each task across multiple candidate budgets K. That design can distinguish retrieval failure from competition failure and test whether routing helps by preserving the right functions, removing the wrong functions, or both.

### 8.4 Pilot-to-Replication Consistency

The v5 pilot produced 2/12=16.67% native success for AgentWeave and 0/12 for all matched baselines. The larger v6 replication produced 6/48=12.5% versus 0/48. Descriptively, the direction is consistent across two non-overlapping frozen samples. The studies are not pooled into a new inferential claim: v5 is treated as hypothesis-generating and v6 as the larger untouched replication.

The sequence matters because repeated tuning on the same benchmark can convert an exploratory result into an apparently strong but non-generalizable claim. Freezing v5, evaluating 48 fresh tasks in v6, and retaining weak historical routing results creates an auditable distinction between development evidence and replication evidence.

### 8.5 Decomposing Where Success Can Arise

End-to-end function-call success can fail at several stages. A required function may be removed during routing; all required functions may survive but close distractors may dominate model selection; the correct function may be chosen with invalid arguments; or execution may fail after a syntactically correct call. The current v6 primary metric intentionally collapses these stages into native BFCL success, while the retention metrics expose only part of the upstream routing behavior.

A deeper evaluation should therefore use a causal-style stage decomposition. Let R denote required-function retention, S valid model selection, A valid argument generation, and E successful execution. End-to-end success requires the conjunction $R \wedge S \wedge A \wedge E$. Reporting each conditional transition - $P(R)$, $P(S|R)$, $P(A|S,R)$, and $P(E|A,S,R)$ - would show whether AgentWeave's benefit comes from candidate construction or whether the downstream model remains the dominant bottleneck.

### 8.6 Result Boundaries and Decision-Relevant Takeaways

Three conclusions are directly supported by the frozen evidence. First, candidate-space construction is experimentally meaningful: the same model and evaluator produce different native outcomes under different presentation strategies. Second, bounded routing materially reduces schema exposure, input tokens, and local inference latency. Third, higher original-candidate retention alone is insufficient to predict native success in this sample. The evidence does not establish state-of-the-art BFCL performance, universal improvement across models, or hardware-independent latency gains.

For system designers, the practical implication is to measure routing and reasoning separately. A production evaluation should report required-tool Recall@K, all-required-tools-retained rate, candidate-set size, schema-token count, routing latency, model latency, native function-call correctness, authorization outcomes, and execution success. This prevents a strong router from being blamed for downstream generation errors and prevents a strong model from masking a router that silently discards required capabilities.

## 9. Reproducibility and Open-Source Artifacts

The AgentWeave repository is designed to make research claims traceable to executable artifacts. The v6 record includes the scored source revision, BFCL/Gorilla revision, workflow run, artifact identifier, and cryptographic artifact digest. Raw and machine-readable results are retained alongside human-readable summaries. The repository also provides reproduction documentation and freezes scored manifests after successful evaluation.

### 9.1 Frozen v6 Record

| Item | Recorded value |
|---|---|
| Study | bfcl-routing-pressure-v6 |
| Tasks | 48 fresh BFCL V4 multiple tasks |
| Model | MadeAgents/Hammer2.1-1.5b |
| BFCL/Gorilla commit | 6ea57973c7a6097fd7c5915698c54c17c5b1b6c8 |
| Scored source head | ca6ff084da4fe5c670421b99e7ad413650e60c33 |
| Workflow run | 31983991285 |
| Artifact | 9275082744 |
| External API spend | $0 |

Table 6. Selected reproducibility metadata retained by the frozen v6 study.

### 9.2 Reproduction Philosophy

A reproduction should preserve the frozen protocol rather than silently update dependencies or rerun against a newer benchmark snapshot. The purpose of pinning is not to claim that one version is permanently authoritative; it is to ensure that the reported numerical result corresponds to a specific executable environment. Future evaluations can and should use newer BFCL versions, but they should receive new study identifiers and be reported as new evidence.

### 9.3 Software and Citation

AgentWeave is released under the Apache-2.0 license and includes citation metadata. Until an archival paper is available, the repository can be cited as software. This manuscript is intended to provide the research narrative connecting the architecture, frozen studies, statistical analysis, limitations, and reproducibility controls. If deposited on arXiv, the corresponding arXiv identifier should be added to the repository citation metadata and future references.

### 9.4 arXiv Submission Considerations

arXiv recommends compact single-spaced manuscripts rather than double-spaced referee mode, requires submitters to inspect the generated PDF during submission, and strongly encourages carefully prepared references with recognizable arXiv identifiers where available. The final arXiv upload should therefore be generated from a clean LaTeX/PDF source package with only files required for compilation, while this Word document serves as an editable manuscript master.

The primary category selected for this work is cs.AI (Artificial Intelligence), because the central contribution concerns agentic orchestration, tool routing, candidate-space control, and AI systems that invoke external capabilities. The paper also has a strong cs.CL connection through language-model function calling, textual tool schemas, and BFCL evaluation; cs.CL is a natural cross-list where permitted by arXiv moderation and endorsement status.

## 10. Limitations, Ethics, and Future Work

### 10.1 Limitations

The strongest limitation is scale. Forty-eight tasks are enough to establish a statistically significant paired difference against zero-success matched baselines in this specific protocol, but they are not enough to characterize broad function-calling performance. Larger untouched replications are required, ideally spanning several BFCL categories and standard benchmark-native data without routing-pressure augmentation.

Second, the study uses one downstream model. Hammer2.1-1.5b is intentionally lightweight and may be unusually sensitive to candidate pressure. Larger models or models trained with different tool-calling objectives may benefit less from pre-inference routing, or may require different routing budgets. A robust conclusion therefore requires a model-by-catalog-size evaluation matrix.

Third, the current router is not a high-recall state of the art retrieval system. Semantic top-8 retains more original candidates. AgentWeave should therefore be compared against stronger dense, sparse, hybrid, and learned reranking baselines. A useful next target is high Recall@K under a strict candidate budget, followed by downstream native evaluation.

### 10.2 Security and Authorization

Reducing model-visible tools can decrease exposure to irrelevant or malicious descriptions, but routing should not be presented as a security boundary. Tool authorization must remain deterministic and fail closed after the model selects an action. Future adversarial evaluation should mix unauthorized, malicious, duplicated, and prompt-injected tools into large catalogs and report unauthorized execution rate, invalid calls, model-visible malicious candidates, and zero-candidate failures.

### 10.3 Environmental and Cost Considerations

The v6 experiment uses a small local model and no paid external inference API. Candidate reduction also lowers the number of input tokens processed by the model in the reported environment. These properties can improve experimental accessibility and may reduce compute cost, but this paper does not claim a lifecycle energy or carbon benefit. Such claims would require direct energy measurement across comparable hardware.

### 10.4 Future Research Agenda

- High-recall hybrid routing: deterministic scope filtering followed by dense/sparse retrieval and reranking.
- Budget curves: evaluate Recall@K, all-required-retained rate, tokens, latency, and native success as K varies.

- Multi-model replication: test small and large open function-calling models plus provider-native tool calling.
- Standard BFCL integration: evaluate an upstream-compatible AgentWeave handler on standard benchmark-native data.
- Adversarial catalogs: study duplicated schemas, malicious descriptions, unauthorized tools, and noisy metadata.
- Catalog scaling: evaluate tens, hundreds, thousands, and eventually enterprise-scale tool sets.
- Learned routing: compare deterministic AgentWeave with supervised and contrastive rerankers trained for function retrieval.

11. Conclusion

This paper argues that function calling should be studied as a two-stage system: candidate-space construction followed by language-model reasoning. AgentWeave implements the first stage through deterministic pre-inference routing while keeping the downstream model fixed. In a frozen 48-task BFCL-derived replication, the routed condition achieves six native successes while three matched alternatives achieve none and simultaneously reduces visible tools, input tokens, and mean local-model latency. The result is preliminary, narrow, and not an official BFCL leaderboard score. Its significance lies in demonstrating that the model-visible function set is itself an important experimental variable. Future work should therefore evaluate routing quality, candidate composition, and downstream function-call behavior jointly rather than assuming that the full source catalog is always the correct model input.

## References

[1] T. Schick, J. Dwivedi-Yu, R. Dessì, R. Raileanu, M. Lomeli, L. Zettlemoyer, N. Cancedda, and T. Scialom. Toolformer: Language Models Can Teach Themselves to Use Tools. arXiv:2302.04761, 2023.

[2] S. Yao, J. Zhao, D. Yu, N. Du, I. Shafran, K. Narasimhan, and Y. Cao. ReAct: Synergizing Reasoning and Acting in Language Models. arXiv:2210.03629; ICLR, 2023.

[3] S. G. Patil, T. Zhang, X. Wang, and J. E. Gonzalez. Gorilla: Large Language Model Connected with Massive APIs. arXiv:2305.15334, 2023.

[4] M. Li, Y. Zhao, B. Yu, F. Song, H. Li, H. Yu, Z. Li, F. Huang, and Y. Li. API-Bank: A Comprehensive Benchmark for Tool-Augmented LLMs. arXiv:2304.08244; EMNLP, 2023.

[5] Y. Qin, S. Liang, Y. Ye, et al. ToolLLM: Facilitating Large Language Models to Master 16000+ Real-World APIs. arXiv:2307.16789; ICLR, 2024.

[6] Q. Lin, M. Wen, Q. Peng, G. Nie, J. Liao, J. Wang, X. Mo, J. Zhou, C. Cheng, Y. Zhao, J. Wang, and W. Zhang. Hammer: Robust Function-Calling for On-Device Language Models via Function Masking. arXiv:2410.04587, 2024.

[7] W. Liu, X. Huang, X. Zeng, et al. ToolACE: Winning the Points of LLM Function Calling. arXiv:2409.00920; ICLR, 2025.

[8] S. G. Patil, H. Mao, F. Yan, C. C.-J. Ji, V. Suresh, I. Stoica, and J. E. Gonzalez. The Berkeley Function Calling Leaderboard (BFCL): From Tool Use to Agentic Evaluation of Large Language Models. Proceedings of the 42nd International Conference on Machine Learning, PMLR 267:48371-48392, 2025.

[9] Berkeley Function Calling Leaderboard V4. Gorilla / UC Berkeley. Current leaderboard and evaluation documentation, 2026.

[10] S. Singla. AgentWeave: open-source software for pre-inference tool routing and multi-agent orchestration. Version 0.6.0, Apache-2.0, 2026. Repository: github.com/sauravsingla/agentweave.

## Reproducibility Statement

The numerical results reported in Sections 5-7 are drawn from the frozen AgentWeave BFCL routing-pressure v6 record and its documented v5 predecessor. The v6 study is immutable after scoring. Any future change to the router, model, sample, distractor construction, or generation protocol should use a new study identifier and fresh evaluation sample. Machine-readable artifacts, workflow metadata, and reproduction instructions are maintained in the public repository.

## Data and Software Availability

AgentWeave source code, benchmark manifests, frozen result summaries, and reproduction documentation are publicly available at github.com/sauravsingla/agentweave under the Apache-2.0 license. BFCL data and evaluation code are maintained by the Gorilla/BFCL project and remain subject to their published repository terms and versioned evaluation procedures.

## Author Statement

This manuscript reports open-source research and benchmark evidence. It does not imply endorsement by BFCL/Gorilla, MadeAgents/Hammer, arXiv, or any employer or affiliated organization. Product and project names are used only to identify public software, models, and benchmarks evaluated or discussed in the paper.

## Appendix A. Evaluation Protocol Details

For each selected BFCL task, the routing-pressure harness first reconstructs the task's original candidate function set and then deterministically augments the model-visible environment to the configured pressure size. Strategy-specific selection is applied only after this common construction. All-tools exposes the complete pressure set; random top-8 uses a fixed deterministic selection procedure; semantic top-8 ranks candidates by semantic relevance; and AgentWeave applies its frozen requirement/capability-aware routing budget. The same downstream Hammer2.1-1.5b model and generation configuration are then used for every strategy. Native BFCL evaluation is performed after restoring the evaluator-facing structure expected by the benchmark.

The principal routing diagnostics are defined as follows. Original-candidate recall is the fraction of the task's original benchmark candidates that remain in the model-visible set after routing. The all-original-candidates-retained indicator equals one only when every original candidate survives. These metrics deliberately do not identify which original candidate is actually required by the natural-language query; they instead provide a conservative diagnostic of how much benchmark information survives compression. Native BFCL success remains the end-to-end function-call outcome under the benchmark evaluator.

A future reproduction should verify five invariants before comparing scores: (1) the exact source and benchmark revisions match the frozen manifest; (2) the 48 task identifiers match the recorded v6 sample and remain disjoint from v5; (3) the model identifier and generation settings are unchanged; (4) strategy budgets and distractor construction match the frozen configuration; and (5) no task-level outcomes are inspected during tuning. If any invariant changes, the resulting experiment should be reported under a new study identifier rather than as a reproduction of v6.

The proposed next experiment should extend this protocol by sweeping the candidate budget K and reporting a routing operating curve. For each K, the evaluation should record required-tool Recall@K when ground-truth function identity is available, all-required-functions-retained rate for multi-function cases, candidate-set compression, prompt tokens, routing latency, model latency, and native BFCL success. This would separate improvements in retrieval quality from improvements caused by simply increasing the number of visible functions, and would enable fair comparison against dense, sparse, hybrid, and learned reranking baselines.

### Appendix A.1. Derived Quantities and Reading the Efficiency Frontier

Several useful quantities can be derived directly from the frozen aggregate measurements without introducing new experimental observations. Relative to all-tools, the AgentWeave condition removes 11.23 visible tools per task on average. The 77,016-token aggregate reduction across 48 tasks corresponds to roughly 1,604 fewer model-input tokens per evaluated task. The mean latency difference of 28.49 seconds per task corresponds to approximately 22.8 minutes less model time across a 48-task pass if the reported mean difference is applied uniformly for descriptive accounting. These derived quantities are not additional measurements; they are arithmetic restatements of the frozen totals and means.

The top-8 baselines help locate AgentWeave on an efficiency frontier. Random top-8 and semantic top-8 both expose exactly eight tools by design, whereas AgentWeave exposes 4.77 on average. Thus AgentWeave uses about 40.4% fewer visible tools than the top-8 conditions. Its aggregate input-token count is about 31.5% lower than random top-8 and 32.0% lower than semantic top-8. Mean model latency is about 10.2% lower than random top-8 and 23.5% lower than semantic top-8. Because these strategies differ in candidate composition as well as size, these contrasts should be read descriptively rather than as isolated causal effects of tool count.

The native-success column makes the frontier especially notable: the most compressed condition is also the only condition with non-zero native success. This does not prove that stronger compression causes better reasoning. A more defensible interpretation is that the particular AgentWeave candidate sets achieved a favorable combination of compactness and decision utility on six tasks. Future budget sweeps are needed to determine whether the relationship is monotonic, U-shaped, or task dependent.

### Appendix A.2. Threats to Validity

Internal validity is strengthened by the matched-task design, fixed downstream model, frozen routing configuration, pinned benchmark revision, and native evaluator. The principal internal threat is that AgentWeave differs from the baselines in more than candidate-set size: its selection rule changes which distractors are visible. This is intentional for the system comparison but means the current study cannot attribute the outcome separately to compression magnitude, capability grouping, requirement extraction, ordering, or other routing details.

Construct validity depends on what the experiment is claimed to measure. Native BFCL success is an appropriate end-to-end function-calling outcome, but original-candidate recall is only a routing diagnostic and is not identical to required-tool Recall@K. The paper therefore avoids treating original-candidate recall as ground-truth retrieval accuracy. A future evaluation with explicit required-function labels should report both required-tool retention and all-required-tools-retained rate.

External validity is deliberately limited. The evaluation uses one small function-calling model, one BFCL category, one pressure size, and 48 fresh tasks. The results may differ for stronger models, longer catalogs, different schema distributions, multi-turn agents, hosted inference, or production authorization constraints. The correct next step is replication across these axes rather than extrapolation from the current percentage reductions.

Statistical conclusion validity is also bounded by sample size. The paired significance result is useful because the same tasks are compared and all six discordances favor AgentWeave, but the confidence interval remains wide. Reporting the interval, absolute success rate, and zero-success baselines together is more informative than presenting the p-value alone.

## Appendix B. Result Interpretation Checklist

This appendix provides a compact checklist for reproducing and interpreting the reported AgentWeave result without extending the empirical claims beyond the frozen evidence.

**Matched comparison integrity.** Confirm that every strategy is evaluated on the same 48 task identifiers, with the same Hammer2.1-1.5b model, generation settings, benchmark revision, and 16-tool pressure construction.

**Primary outcome.** Treat native BFCL success as the end-to-end scored outcome. Do not substitute routing recall, tool count, or a synthetic proxy for native evaluator acceptance.

**Effect-size interpretation.** Report the observed +12.5 percentage-point paired difference together with the 95% bootstrap interval (+4.17 to +22.92 points) and exact McNemar p=0.03125; retain the absolute 6/48 success rate beside the relative comparison.

**Efficiency interpretation.** Report tool, token, and latency reductions as measurements of this local frozen setup. Do not extrapolate the 50.95% latency reduction to different hardware or serving systems without new measurements.

**Retention interpretation.** Preserve the negative result that semantic top-8 has stronger original-candidate retention than AgentWeave but zero native successes. This prevents candidate recall from being misrepresented as the sole explanation.

**Replication boundary.** Keep v5 and v6 distinguishable. The 12-task pilot and 48-task replication are directionally consistent but should not be silently pooled into a stronger statistical claim.

**Benchmark boundary.** Describe the experiment as BFCL-derived routing pressure, not an official full BFCL leaderboard submission. The original questions and native evaluator are preserved, while the model-visible candidate environment is experimentally controlled.

**Causal boundary.** The current evidence shows that candidate presentation changes outcomes under a fixed model. It does not identify which AgentWeave routing feature is individually causal; ablation and candidate-composition studies are required.

**Next decisive experiment.** Sweep candidate budget K on a fresh holdout, record required-tool Recall@K and all-required retention, measure distractor similarity, and evaluate multiple downstream models. This would test whether the observed advantage persists across operating points and model capacities.